# The Future of Neural Networks

Sachin Lakra[1], T.V. Prasad[2] and G. Ramakrishna[3]
[1]Research Scholar, CSE, KL University, Guntur, Andhra Prdesh & Associate Professor, IT, Manav Rachna College of Engineering, Faridabad, Haryana (sachinlakra@yahoo.co.in)
[2]Dean(Academics), Lingaya's University, Faridabad, Haryana (tvprasad2002@yahoo.com)
[3]Professor, CSE, K L University, Guntur, Andhra Pradesh (ramakrishna_10@yahoo.com)

**ABSTRACT**
*The paper describes some recent developments in neural networks and discusses the applicability of neural networks in the development of a machine that mimics the human brain. The paper mentions a new architecture, the pulsed neural network that is being considered as the next generation of neural networks. The paper also explores the use of memristors in the development of a brain-like computer called the MoNETA. A new model, multi/infinite dimensional neural networks, are a recent development in the area of advanced neural networks. The paper concludes that the need of neural networks in the development of human-like technology is essential and may be non-expendable for it.*

**KEYWORDS**
Neural Network, Pulsed Neural Network, Neuro-fuzzy Computing, MoNETA.

**INTRODUCTION**
Neural networks are capable of handling a wide variety of the type of problems which involve finding trends in large quantities of data. They are better suited than traditional computer architecture to handling problems that humans are naturally good at and which computers are not able to handle very well. For example, pattern recognition, generalizing, clustering, etc., are problems which humans handle better than computers and so do neural networks. Research is also focused in the direction of developing new models of neural networks that are better at solving these problems.

Certain problems remain to be solved. These problems are described by the following questions:
- Will neural networks ever fully simulate the human brain?
- Will neural networks be as complex and as functional as the human brain?
- Will a machine ever be conscious of its own existence?

**CHARACTERISTICS OF NEURAL NETWORKS**
According to [1] the theoretical understanding of how cognition arises in the brain has been advanced by the understanding of artificial neural networks that display
- learning from experience,
- perceptual discrimination of inarticulable features,
- development of a hierarchy of categories or framework of concepts,
- spontaneous inductive inference in accordance with past experience ("vector completion").
- Sensorimotor coordination between sensory inputs and motor outputs
- short term memory with information selective decay time
- variable focus of attention.

**GENERATIONS OF NEURAL NETWORKS**
Wolfgang Maass [2] has identified three generations of past and current artificial neural network research and makes the following observations.

The first generation has the McCulloch-Pitts neuron (also known as a perceptron or a threshold-gate) as the basic unit of computation. Models of the first generation, such as the multi-layer perceptron, use digital input and output, either binary or bipolar. A Boolean function can be computed by using a multi-layer perceptron with a single hidden layer.

The second generation has as its basis computation units (neurons) that use an activation function of a continuous set of possible output values. Commonly, these activation functions are the sigmoid, or the hyperbolic tangent. Second generation neural networks can also compute any boolean function by using a threshold. Second generation network are capable of computing certain boolean functions with fewer neurons than first generation neural networks. Further, second generation networks with one hidden layer can approximate any continuous, analog function. Also, second generation networks support learning algorithms based on gradient descent, such as error back-propagation.

The third generation of artificial neural networks is based on spiking neurons, or "integrate and fire" neurons. These neurons are based on recent insights from neurophysiology, specifically the use of temporal coding to pass information between neurons. These networks, like those of the second generation, can approximate continuous functions well by using temporally encoded inputs and outputs [2, 3]. Further, there are functions that require even fewer neurons in a pulsed neural net to approximate than would be needed in a second generation network [2].

All three of these generations are simplifications of what is known about the physiology of biological neurons but the third generation is the model with the highest fidelity.

**NEURAL NETWORKS AND CONSCIOUSNESS [4]**
Simulating human consciousness and emotions are in the realm of science fiction. There are inconclusive philosophical arguments about what consciousness is, and if it can possibly be simulated by a machine. The question is that assuming that the physicality of souls is non-existent, how can human beings make the jump from, as one researcher puts it, "an electrical reaction in the brain to suddenly seeing the world around one with all its distances, its colors and chiaroscuro?". One attempt at resolving these issues has been presented at [5].

Current models of information processing inside the central nervous system describe it as occurring through hierarchically organized and interconnected neural networks. For instance, information captured by the eyes is firstly hierarchically processed at the level of the retina (from the photoreceptory rods and cones, to the ganglion cells), followed by hierarchical processing within the levels of primary, secondary, and tertiary sensory and interpretatory cortical regions (all of them being additionally constituted of hierarchies of several neural networks) [6]. Interconnections within neural networks and between the neighboring neural networks in this hierarchy are achieved through synapses (one neuron having approximately 40000 synaptic connections with neighbors). During learning, a significant role is played by brainwaves in distribution and memorizing of hierarchically processed information throughout the whole cortex[7].

Along with the development of experimental techniques enabling physiological investigation of interactions of hierarchically interconnected neighboring levels of biological neural networks, significant contribution in establishing the neural network paradigm was given by theoretical breakthroughs in this field during the past decade [8].

Computers today are much faster than a neuron in the brain but the brain is far superior than a computer at handling certain complex tasks, such as the image processing and recognition, orientation and movement in the space of changeable characteristics, speech recognition, etc. The reason for this superiority of the brain is due to parallel information processing.

Neural networks have many properties suitable for being used to replicate the functioning of the brain: parallel functioning, relatively quick solving of complicated tasks, distributed information, weak sensitivity on local damages, as well as learning abilities, i.e. adaptation to changes in environment, and improvement based on experience.

Due to these properties, a sufficiently big network, with adequate training could accomplish an arbitrary task, without knowing a detailed mathematical algorithm of the problem. However, today the only solution available is find a suitable network topology and training rules for every particular task [8].

Perhaps neural networks can provide some insight into some of the easier problems of consciousness: how does the brain process environmental stimulation? How does it integrate information? But, the real question is, why and how is all of this processing, in humans, accompanied by an experienced inner life, and can a machine achieve such a self-awareness? However, the whole future of neural networks does not reside in attempts to simulate consciousness. Another issue is of how to improve the systems that exist.

**RECENT ADVANCES AND FUTURE APPLICATIONS OF NEURAL NETWORKS [4]**
Some of the recent advances and future applications of neural networks are mentioned below.

**Integration of fuzzy logic into neural networks [4]**
Fuzzy logic and neural networks have been integrated for uses as diverse as automotive engineering, applicant screening for jobs, the control of a crane, and the analysis of an ECG.

The result of integrating fuzzy logic into neural networks is the neuro-fuzzy inference system(NFIS) that in addition to the processing steps of a neural network, first fuzzifies crisp input data, uses fuzzy rules to process this data to obtain a fuzzy result and then defuzzifies this fuzzy result to obtain a crisp final output. Examples of NFIS's are Adaptive NFIS (ANFIS), Co-Active NFIS (CANFIS), Modular Adaptive NFIS (MANFIS), etc. Applications of neuro-fuzzy computing are diverse as mentioned at [15-17].

**Pulsed or spiking neural networks**
Most practical applications of artificial neural networks are based on a computational model involving the propagation of continuous variables from one processing unit to the next. In recent years, data from neurobiological experiments have made it increasingly clear that biological neural networks, which communicate through pulses, use the timing of the pulses to transmit information and perform computation. This realization has stimulated significant research on pulsed neural networks, including theoretical analyses and model development, neurobiological modeling, and hardware implementation [9].

Traditionally, it was believed that neurons communicated information in their mean firing rate. In other words, one neuron would receive information from another by "counting" the number of spikes from that neuron over some extended period of time and determining the mean time between firings. Specifically, a shorter time period implies a higher activation. This type of information coding is known as *rate coding* and has been one of the dominant tools for measuring neuron activity over the past 80 years of neurological study [10]. Correspondingly, a majority of artificial neural network models have used rate coding, in the form of real number values representing an activation level.

Recently, new discoveries and advances in neurophysiology have encouraged new explorations in alternative information coding schemes in artificial neural networks. Neurons have been found in the primate brain that responds selectively to complex visual stimuli after as few as 100-150 ms after the stimulus was presented. This information must have passed through approximately 10 layers of processing between the initial photoreceptors and the neurons that selectively respond to the stimuli.

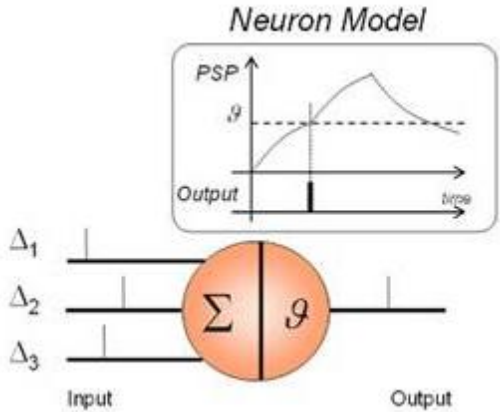

Figure 1. Basic Model of a Pulsed or Spiking Neuron[25].

Given this, it was argued that each individual processing stage would have only 10 ms to complete and that this amount of time is insufficient for rate coding as a means of information passing [11].

Another information coding scheme must therefore be at work in the brain, one that may be useful in artificial neural networks. Codes based on the temporal relationship between the firing of neurons are a promising alternative.

Using such codes it is possible to transmit a large amount of data with only a few spikes, as few as one or zero for each neuron involved in the specific processing task [11].

**Hardware specialized for neural networks [4]**

Some neural networks have been hardcoded into chips or analog devices. The primary benefit of directly encoding neural networks onto chips or specialized analog devices is speed. NN hardware is used in a few specialized areas, such as those areas where very high performance is required (e.g. high energy physics) and in embedded applications of simple, hardwired networks (e.g. voice recognition).

When NN algorithms develop to the point where useful things can be done with 1000's of neurons and 10000's of synapses, high performance NN hardware will become essential for practical operation. [12]

*The Modular Neural Exploring Travelling Agent (MoNETA) [13]*

The MoNETA (Modular Neural Exploring Traveling Agent) is the software being designed at the Department of Cognitive and Neural Systems, Boston University, which will run on a brain-inspired microprocessor under development at HP Labs, California. This microprocessor will function according to the fundamentals that demarcate mammals most profoundly from machines. MoNETA will be capable of performing functions that existing or past computers never were capable of. It will recognize its environment, filter out useless information, integrate the useful information into the evolving structure of its virtual reality, and in some situations, devise plans that will guarantee its survival.

Researchers had doubted for a long time that artificial intelligence can not be implemented using the traditional von Neumann hardware architecture, with Boolean logic at its core and physical separation between memory and processing. This observation led to the development of a new type of electronic devices called memristors by Hewlett-Packard. Before this device, it was not feasible to develop hardware with the form factor of a mammalian brain, the low power requirements, and the near-instantaneous internal communications. These three functionalities form the key to developing anything that resembles the brain and thus can be trained to behave and function like a brain. Basically, memristors are sufficiently small, cheap, and efficient to be suitable for implementing these three key functions. More importantly, these devices have characteristics that resemble those of synapses.

The research that will yield this artificial intelligence capable hardware architecture is being carried out at the U.S. Defense Advanced Research Projects Agency (DARPA). When the work on this brain-inspired microprocessor is complete, its first role will likely be in the U.S. military, in the form of MoNETA being used as replacements for humans in scout vehicles identifying roadside bombs and helping to defuse them, or navigating hostile terrain.

However, every piece of AI is specialized and must be programmed specifically to carry out its task. What is needed is a general-purpose intelligence that can solve and handle any problem; one that is adaptive and can handle any new environment without the requirement of constant retraining as in the case of a neural network.

Consider the capabilities of the MoNETA-enabled military scout vehicle. It will be capable of going into a mission with partially known objectives that may get altered unexpectedly. It will be able to negotiate previously unknown stretches of terrain, recognize a pattern that may signify hostile activity, formulate a plan, and maneuver itself out of the hostile region. If the path is blocked, it will be able to make an instantaneous decision and move off the road to reach its destination. Intuition, pattern recognition, improvisation, and the ability to negotiate ambiguity: All of these things are done really well by the human brain.

Certain tasks that a number-crunching computer can not do can be carried out very easily by a rat. The architecture of the rat's brain is what makes the rat's brain superior. An adult rat's brain consists of 21 million nerve cells called neurons (the human brain has about 100 billion). Neurons communicate with each other via dendrites and axons, transmitting electrical impulses from one neuron to another. Most of the processing performed in the nervous system occurs in the junctions between neurons. This junction is a space called a synapse and lies between one neuron's dendrite and a neighboring neuron's axon.

Computational neuroscience till date has concentrated its efforts on developing software that can simulate or replicate a mammal's brain using the von Neumann architecture. This architecture separates the processor where data is processed from the memory where it is stored, and it has been the fundamental basis of computer architectures since the 1960s.

Researchers thought that, given a sufficient number of powerful CPUs, building programs that emulate the functionalities of the brain will be a logical outcome. However, saying this is equivalent to stating that given a sufficient number of words, creating a book is the logical outcome.

Architecture is the key to imbibing natural intelligence capabilities in a computer. To understand why, compare the path of a hypothetical bit of data inside a conventional microprocessor with its path inside a brain. In a von Neumann architecture based computer, the memory and processor are separated by a data channel, called the data bus, between the area where the data is stored and where it is processed. The data bus has a fixed capacity for transmitting data implying that only limited quantities of data can be communicated and processed at a given instant. The processor reserves a small number of memory areas, called registers, for storing data during computation. After completing the required computation, the processor writes the results back to memory, again, using the data bus. With small computation requirements the data bus is efficient and effective. To reduce the amount of data being transmitted on the fixed-capacity data bus, most modern processors enhance the registers with a cache memory that temporarily stores data very close to the point of computation. If a computation that is carried out repeatedly requires multiple pieces of data, the processor will keep them in the cache, which the processing unit can then access much more efficiently than it can access the main memory.

However, this entire architecture consisting of the registers and cache will not work for handling the computational challenges met with while trying to simulate a human brain. Even a simple brain has an extremely large number of neurons connected by a many times larger number of synapses. Therefore, to simulate such a large structure will require a cache as large as the computer's main memory, which would make the machine being built useless.

Scientists call this brain-simulating system a neuromorphic architecture. A large percentage of the computing work and power of such a system is used up in attempting to replicate the signal processing that happens inside the brain's synapses. To model a single synapse requires the following to happen in the machinery: The synapse's state, that is, how likely it is to pass on an input from a neuron (the major factor in how strong the association is between any two neurons) resides in a location in main memory. To alter that state, the processor must package an electronic signal for transfer over the data bus. That signal must travel between 2 to 10 centimeters to reach the main memory and then must be translated to actually access the required memory location.

Multiply this sequence of steps by up to 8000 synapses, which is the average number of synapses found in a single rat neuron. Next, multiply this number by the number of neurons in the brain, which is in billions. This is the model of one *millisecond* of brain activity.

A biological brain is able to quickly execute this massive flash of simultaneous information processing in a small amount of time, since it has a number of shortcuts. The following is the set of steps that is executed in a brain: Neuron 1 sends out an impulse, which contains information, down the axon to the synapse of the destination, that is, Neuron 2. The synapse of Neuron 2, which has its own state stored locally, assesses the importance of the information coming from Neuron 1 by integrating it with its own previously stored state and the strength of its connection to Neuron 1. Then, the information from Neuron 1 and the state of Neuron 2's synapse flow toward the body of Neuron 2 over the dendrites. During the transmission of these two pieces of information to the body of Neuron 2, processing of this information occurs and only a single processed value reaches Neuron 2. There is no need for the brain to extract information from one neuron, process it, and then resend it to a different group of neurons. Instead, in the mammalian brain, storage and processing are carried out at the same time and in the same place.

Reproducing the brain's functionality on even the most advanced supercomputers would require a dedicated power plant, whereas brains can operate at around 100 millivolts. CMOS logic circuits require a much higher voltage of close to 1 volt to function properly, and the higher operating voltage means that more power is expended in transmitting the signal over wires as a result of the von Neumann architecture.

DARPA is attempting to change the von Neumann architecture to merge memory and processor. The memristor is the best technology for the task because it is the first memory technology with enough power efficiency and density to rival biological computation structures.

A memristor is a two-terminal device whose resistance changes depending on the amount, direction, and duration of voltage that is applied to it. Whatever its past state, or resistance, it freezes that state until another voltage is applied to change it; maintaining that state requires no power. This is different from a DRAM cell, which requires regular charge to maintain its state. However, the memristive state of the device does decay over time. That decay can occur in hours or centuries depending on the material, and stability must often be traded for energy requirements. In attempting to reduce energy requirements, stability is hampered. This is why the memristor has not been perfected yet.

This architecture is similar to the neural network architecture except that the neurons will be replaced by memristors and the computation will be performed by hardware.

**Multi/Infinite Dimensional Neural Networks**

A new model of neural networks called Multi/Infinite Dimensional Neural Networks(MDNN), which are a generalization of One-Dimensional Neural Networks have been developed in theory [18]. The theory is based on the concept of generalizing one-dimensional logic gates to multi-dimensional logic gates and in a similar manner one-dimensional neural networks have been generalized to MDNN's[19]. An MDNN takes inputs from the states of a multi-dimensional hypercube and learns stable states after processing. The architecture of an MDNN is represented by a Tensor State Space Representation [20]. Tensor products are utilized to calculate the output of each neuron in an MDNN. An MDNN uses a back propagation algorithm generalized to a complex valued neural network and is based on complex signum and complex sigmoid functions [21, 22].

This generalization of neural networks has been applied to the development of a unified theory of control, communication and coding[18], the three concepts central to cybernetics. Applications of MDNN's are in the areas of optimal binary filters[23] and complex valued neural associative memory on the complex hypercube[24].

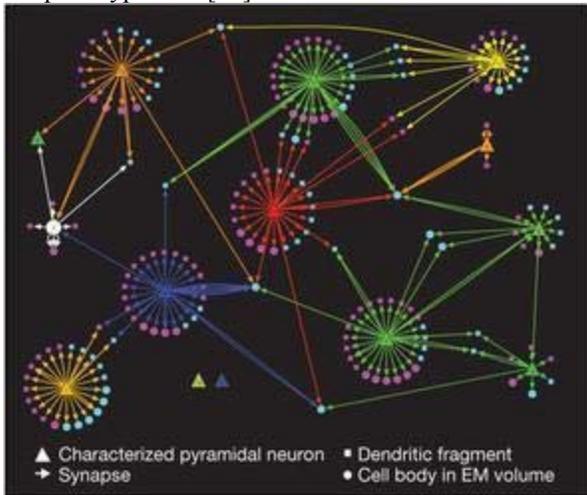

Figure 2. A 3-dimensional neural network from the human brain[26].

**IMPROVEMENT OF EXISTING TECHNOLOGIES**
All current NN technologies will most likely be vastly improved upon in the future. Everything from handwriting and speech recognition to stock market prediction will become more sophisticated as researchers develop better training methods and network architectures. Neural Networks might, in the future, allow:
- robots that can see, feel, and predict the world around them
- improved stock prediction
- common usage of self-driving cars
- composition of music
- handwritten documents to be automatically transformed into formatted word processing documents
- trends found in the human genome to aid in the understanding of the data compiled by the Human Genome Project
- self-diagnosis of medical problems using neural networks[4].

**A WORD OF CAUTION**
Although neural networks do seem to be able to solve many problems, they are not magic. Overconfidence in neural networks can result in costly mistakes. As an example [14] the Pentagon, USA required that the tanks in the US army be made harder to attack. As part of this attempt, the research team decided to fit each tank with a digital camera hooked up to a computer. The computer would continually scan the environment outside for possible threats (such as an enemy tank hiding behind a tree), and alert the tank crew to anything suspicious. Computers are really good at doing repetitive tasks without taking a break, but they are generally bad at interpreting images. The only possible way to solve the problem was to employ a neural network.

A hundred images were taken of tanks hiding behind trees and another 100 of trees without a tank hiding behind them. Out of these 200 images, 100 were fed into a mainframe computer and the neural network was asked to identify whether there was a tank in the image or not. Mistakes made by the neural network were corrected on the basis of feedback given to it by the scientists working on the project. The neural network finally learned how to identify an image with a tank hiding behind the trees. To check the effectiveness of this ability of the neural network, the other 100 images were also fed into the neural network as test data and the network identified them correctly. The neural network was then subjected to an independent set of images from another source. However, the results were random. The explanation that came out for the system's success on the first 200 photographs was that out of the first 200 photographs, the ones with tanks behind the trees were taken on a sunny day, whereas the others were taken on a cloudy day. The neural network was distinguishing the two types of photographs correctly only in that was identifying whether there was cloud cover or not.

**CONCLUSION**
It is clear that the use of neural networks is going to increase in terms of both the number and type of their applications in the future. The integration of neural networks with fuzzy logic is already delivering good results towards more effective and efficient solutions to problems otherwise better handled by human beings. The future of neural networks is bright and current research seems to be moving in the right direction towards the ultimate goal of all artificial intelligence, namely, the development of a humanoid robot that can work and think like a human being.

**FUTURE SCOPE**
An understanding of the future of neural networks and their applications will help researchers to appreciate the importance and essentiality of their role in the development of a human-like artificial brain. Artificial Intelligence research is directed towards the development of artificial life which is not possible to be created without an artificial brain-like processing unit forming an integral part of its behavioural and physical structure.